\documentclass[lettersize,journal]{IEEEtran}
\usepackage{amsmath,amsfonts}
\usepackage{algorithmic}
\usepackage{algorithm}
\usepackage{array}
\usepackage[caption=false,font=normalsize,labelfont=sf,textfont=sf]{subfig}
\usepackage{textcomp}
\usepackage{stfloats}
\usepackage{url}
\usepackage{verbatim}
\usepackage{graphicx}
\usepackage{cite}
\usepackage{multirow}
\usepackage[section]{placeins}
\begin{document}

\title{Residual Chain Prediction for Autonomous Driving Path Planning}

\author{Liguo Zhou$^{*}$, Yirui Zhou$^{*}$, Huaming Liu, Alois Knoll
\thanks{Liguo Zhou, Yirui Zhou, and Alois Knoll are with the Chair of Robotics, Artificial Intelligence and Real-time Systems, Technical University of Munich}
\thanks{Liguo Zhou is with the School of Computer Science, Nanjing University of Information Science \& Technology}
\thanks{Huaming Liu is with the School of Computer and Information Engineering, Fuyang Normal University}
\thanks{* These authors contributed equally to this work.}
}

\markboth{Manuscript}%
{Shell \MakeLowercase{\textit{et al.}}: A Sample Article Using IEEEtran.cls for IEEE Journals}


\maketitle

\begin{abstract}
In the rapidly evolving field of autonomous driving systems, the refinement of path planning algorithms is paramount for navigating vehicles through dynamic environments, particularly in complex urban scenarios. Traditional path planning algorithms, which are heavily reliant on static rules and manually defined parameters, often fall short in such contexts, highlighting the need for more adaptive, learning-based approaches. Among these, behavior cloning emerges as a noteworthy strategy for its simplicity and efficiency, especially within the realm of end-to-end path planning. However, behavior cloning faces challenges, such as covariate shift when employing traditional Manhattan distance as the metric. Addressing this, our study introduces the novel concept of Residual Chain Loss. Residual Chain Loss dynamically adjusts the loss calculation process to enhance the temporal dependency and accuracy of predicted path points, significantly improving the model's performance without additional computational overhead. Through testing on the nuScenes dataset, we underscore the method's substantial advancements in addressing covariate shift, facilitating dynamic loss adjustments, and ensuring seamless integration with end-to-end path planning frameworks. Our findings highlight the potential of Residual Chain Loss to revolutionize planning component of autonomous driving systems, marking a significant step forward in the quest for level 5 autonomous driving system.
\end{abstract}


\section{Introduction}
\IEEEPARstart{A}{utonomous} driving technologies have significantly advanced the quest for fully autonomous vehicles, revolutionizing transportation systems and mobility. Traditional autonomous driving systems utilize a modular deployment strategy, categorized into three core functionalities: perception, prediction, and planning~\cite{Teng_2023}. Which are called Modular Pipelines. At the heart of modular pipelines is the pursuit of developing reliable and efficient path planning algorithms, which are crucial for ensuring that vehicles navigate through complex environments safely.

Historically, path planning methodologies have relied on intricate rule-based designs such as Dijkstra and A* Search Algorithm~\cite{ASTAR} or sampling-based method such as Rapidly exploring Random Tree (RRT) and Probabilistic Road Map (PRM). Despite their effectiveness in static environments, often fall short in dynamic and unpredictable urban situations, due to the fact that they rely on manually defined costs and reward functions that could lead to optimal driving behaviors. Accordingly, there is a discernible shift in the field towards leveraging neural networks as a foundational basis for the evolution of path planning strategies, with a particular emphasis on the accelerated advancement of end-to-end path planning methodologies. These learning-based approaches promise improvements in adaptability and performance with the acquisition of larger dataset~\cite{hu2022stp3, hu2023planningoriented,vitelli2021safetynet}. End-to-end path learning simplifies the design and development process of the whole system. It reduces the need for manually designed features extraction and preprocessing steps, as the model can learn to extract useful features directly from the data. Learning-based path planning methods have the potential to significantly enhance AD system's ability to understand and interact with its environment. 

In the realm of end-to-end path planning, a sequence of future trajectory points is deduced from either raw sensor data or preprocessed data acquired from the vehicle sensors. Conventionally, predicted path points facilitate a direct comparison with expert navigated path points, utilizing distance metrics such as the Manhattan distance (L1 norm) or Euclidean distance (L2 norm), commonly employed within the scope of imitation learning. This methodology, however, presents notable challenges, including difficulties in model convergence and loss in generalization capability. To mitigate these issues, a transition from global to relative coordinates—measured in relation to the initial point—is proposed. Furthermore, additional methodologies such as the Frechet Distance, Chamfer Distance, and Hausdorff Distance, provide alternative means of calculating the discrepancies between predicted and expert-navigated path points. While these methods offer unique perspectives in measuring discrepancies, they fall short in addressing the crucial component of temporal dependencies inherent in dynamic driving environments. Dynamic Time Warping (DTW) presents a methodology for calculating the discrepancy between two temporally misaligned sequences, embodying an approach that contemplates the entire sequence of data points~\cite{dtw}. However, DTW lacks in its provision for adjustments to future points predicated on their pre-predicted path points. These methods also face the problem of covariate shift~\cite{ross2011reduction}.

To rectify this shortfall, we propose the innovative concept of \textbf{Residual Chain Loss}. This approach is designed to enhance both the accuracy and robustness of autonomous driving path planning by introducing a mechanism for dynamic loss adjustment without increasing calculation cost. This adaptation ensures an incorporation with a temporal dimension that reflects the sequential nature of driving decisions, closely mirroring the complex dynamics of real-world driving scenarios.
To sum up, the key contribution of this work include:
\begin{itemize}
    \item \textbf{Enhanced Handling of Covariate Shift:} By adopting relative coordinates as outputs, our method facilitates a more effective alignment between expert and predicted points within the framework of imitation learning. This strategic choice also offers a robust solution to the problem of covariate shift. 
    \item \textbf{Dynamic Loss Adjustment:} Introduction of a dynamic loss adjustment mechanism enables the model to adapt its learning process based on the sequence of previously predicted points, enhancing model accuracy over time without increasing much computation time.
    \item \textbf{Compatibility with End-to-End Path Planning Models:} Our approach is designed to be compatible with a variety of learning-based models, facilitating its adoption and integration into a broad spectrum of path planning model.
\end{itemize}

\section{Related Work}
\noindent\textbf{Trajectory-Based Methods.} Trajectory-based Methods could be classified as Traditional algorithm of map-based path planning algorithms and sampling-based path planning algorithm. Most commonly used traditional algorithms could be defined as A-Class~\cite{overview}. Introduced by E.W. Dijkstra in 1959~\cite{dijkstra1959note}, Dijkstra Algorithm employs a greedy approach to iteratively explore as many subnodes as possible, subsequently utilizing the technique of relaxation to refine and optimize the choice of paths. A* Algorithm is the most commonly used method, improved on Dijsktra Algorithm, A* Algorithm introduces the heuristic function, which makes the algorithm more efficient than Dijkstra Algorithm. Most commonly applied sampling-based methods are Rapidly exploring Random Tree (RRT)~\cite{rrt} and Probabilistic Road Map (PRM)~\cite{rpm}. These algorithms rely on human engineering rather than on data to achieve functionalities. As a result, while they exhibit robust performance in static environments, their ability to navigate complex dynamic environments remains limited.

\noindent\textbf{End-to-end Path Planning.} End-to-end path planning represents a paradigm shift from traditional modular pipelines, which distinctly segregate perception, prediction, and planning modules. This holistic approach not only simplifies the integration of these key components but also facilitates comprehensive optimization, potentially reducing computational costs~\cite{chen2023endtoend}. End-to-end path planning methods could be classified as two streams: Imitation Learning and Reinforcement Learning. Within the domain of Imitation Learning, two subcategories emerge, namely Behavior Cloning (BC)~\cite{bc} and Reverse Reinforcement Learning (RIL)~\cite{Snoswell_2020}. Contrary to Behavior Cloning (BC), which directly mimics expert demonstrations to predict actions, Inverse Reinforcement Learning (IRL) deduces the underlying reward function from expert demonstrations. This approach divides into two key tasks: deriving the cost function through end-to-end learning and optimizing trajectory sampling efficiently~\cite{chen2023endtoend}. IRL offers a deeper insight into expert decisions by understanding their motivations. Reinforcement Learning is the learning of trial. It requires more data in comparison with supervised learning, thus almost all ADS based on RL are investigated in simulation~\cite{chen2023endtoend}. Up to now, RL has successfully work on a car driving on an empty lane on the street, but IL has done it 3 decade before. In the field of end-to-end path planning, RL still has a long way chasing after IL. Recently, there are some works combining supervised learning with RL. Implicit affordances~\cite{toromanoff2020endtoend} and GRI~\cite{chekroun2022gri} pretrain CNN encoder using auxilary tasks such as object detection, semantic segmentation and classification. The image encoder is frozen, policy is trained on implicit affordance of outputs of the frozen image encoder~\cite{toromanoff2019deep}.

\noindent\textbf{Imitation Learning (IL) and Behavior Cloning (BC).} Initially, Behavior Cloning was applied to end-to-end path planning in ~\cite{bojarski2016end}, setting a precedent for subsequent advancements. Progressions in this field have seen the incorporation of multi-sensor fusion ~\cite{chen2022learning}, auxiliary tasks ~\cite{codevilla2019exploring}, and improved expert-design ~\cite{zhang2021endtoend}, each contributing to the enhanced performance of BC in the realm of autonomous navigation. However, despite its considerable successes across a range of applications, BC encounters intrinsic limitations, particularly when faced with the challenge of capturing long-tail sequences that are prevalent in real-world scenarios~\cite{scheel2021urban}. The algorithm has already applied on roads~\cite{hawke2019urban}, including urban driving and highways. In our work, we employ an imitation learning approach, sampling specific points along with their past and future points and their position information. By treating the future points as the expert, we fit the predicted path points based on past points to these future positions. This method allows us to leverage the accuracy of future data as a guiding benchmark for training our model to predict the trajectory more accurately, bridging the gap between past observations and future states. 

\noindent\textbf{Distance Metrics.} In the realms of imitation learning and behavior cloning, our primary objective centers on refining the model's outputs to more closely align with the expert's behavior. This endeavor becomes particularly salient within the purview of path planning, where the choice of a distance function between ground truth and predicted path points emerges as a pivotal consideration. Traditional methodologies employ Manhattan and Euclidean distances, yet alternative metrics such as Frechet, Chamfer, and Hausdorff distances present nuanced variations for evaluation. Furthermore, Dynamic Time Warping (DTW)~\cite{dtw} offers an advanced framework for assessing the distance between two temporally misaligned sequences. These conventional approaches applied in BC, predicated upon deriving full future path point predictions from a given point and its past points, are prone to incremental deviations from the expert’s path. BC treats each points independently~\cite{chen2023endtoend}. Thereby necessitating the covariate shift~\cite{ross2011reduction}, several on-policy methods have been proposed to address this issue~\cite{JMLR:v15:judah14a}~\cite{ross2014reinforcement}. Other than that, Data as Demonstrator (DAD)~\cite{Arun2014DATAAD} proposes an innovative solution by pairing predicted point with the next expert point in the sequence. We depart from the conventional paradigm of predicting the coordinates of future points. Instead, our outputs are formulated as relative coordinates between adjacent path points in the sequence. This adjustment is accomplished by modifying the ground truth to represent the Manhattan distance between the expert coordinates and the sum of the initial point coordinates and the cumulative relative coordinates between adjacent pre-predicted points in the sequence, which are the pre-predicted outputs. Such a loss function meticulously accounts for the dynamic adjustments predicated on previous predictions while simultaneously assessing the relative displacement between predicted and expert points. 
\section{Method}

\subsection{End-to-End Driving Model}

The driving process of a human driver can be described as Equation~\ref{drive}, where $X$ is what the driver sees and hears, $Y$ is the driver's operations to the vehicle, and $f$ is the driver's nervous system that has been trained in driving school and past driving experience. If we are passengers, we can see $X$ and $Y$, but we can't see the processing in $f$ because that's in the driver's nervous system.
\begin{equation}
Y = f(X).
\label{drive}
\end{equation}

Similar to human drivers, the end-to-end machine learning model of autonomous driving has two ends for inputting $X$ and outputting $Y$, and all the other processes are done unexplainably inside the model $f$, such as the feed-forward of the deep neural network. In the extreme case, the input $X$ of the end-to-end model is the raw data of the sensors, and the output $Y$ is the operating commands to the vehicle. In most cases, the input $X$ is preprocessed sensor data, and the output $Y$ is a planned path in the future. The planned path, which is denoted by a series of positions, is then converted into vehicle control commands through the control algorithms.

\begin{figure*}[htb]%
    \centering%
    \includegraphics[width=\textwidth]{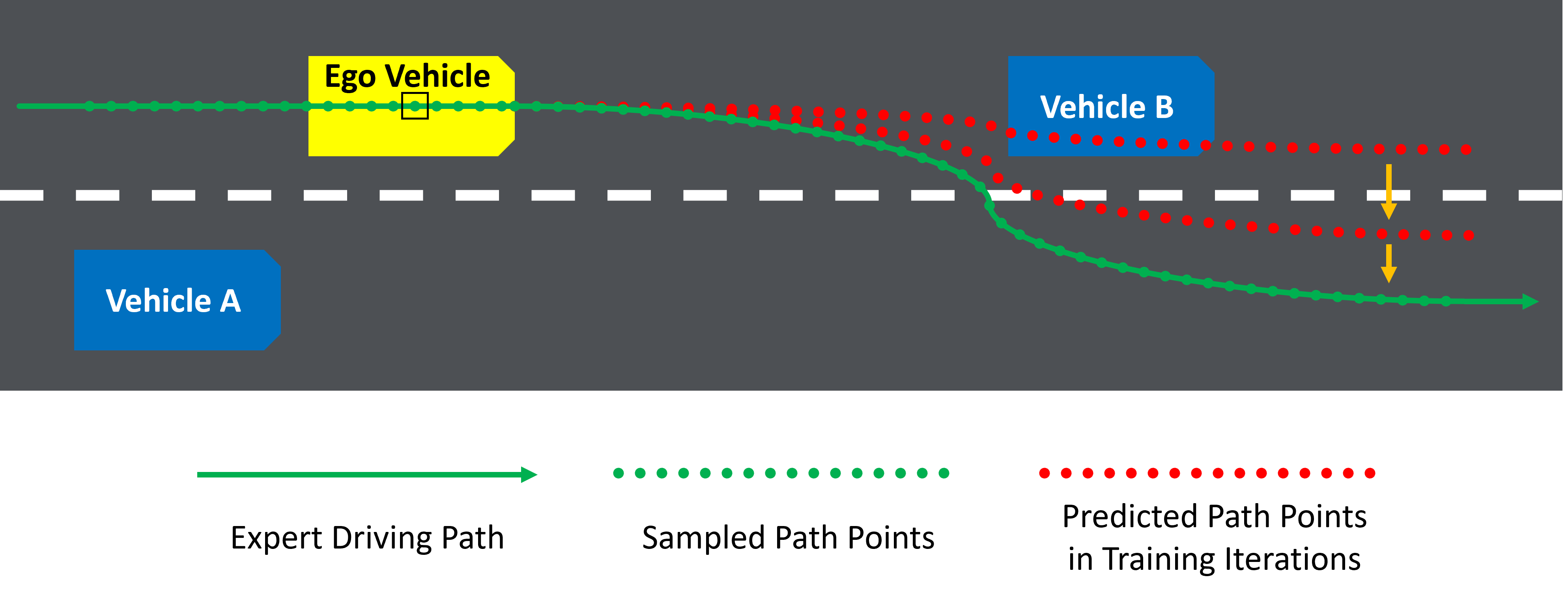}%
    \caption{Learning a Path by Imitation.}%
    \label{fig:il}%
\end{figure*}%

\subsection{Path Planning based on Imitation Learning}

Figure~\ref{fig:il} shows the process of learning a path by imitation:

\begin{itemize}
\item[1] 

Sampling historical driving paths at equal time intervals. Each sampled path point includes the vehicle's position and rotation, which denotes the vehicle pose. Each point also has corresponding environment data, such as images and point clouds, and vehicle status, such as throttle, wheel angle, and braking status. 

\item[2]

Select a sampled point as the current point and split the other sampled points into past and future. The point in the "$\square$" denotes the current pose. The points before it belong to the past, and the points behind it belong to the future. The current point and $n$ consecutive points before it, as well as their corresponding environment data and vehicle statuses, are the input of the path planning model. The $m$ consecutive points after the current point are the ground truth of the predicted path points of the model.

\item[3]

Fitting the predicted path points to the ground truth. At the beginning of training the path planning model, the predicted path points are random values. In the training iterations, the predicted path points become closer and closer to the ground truth by optimizing the model parameters.

\end{itemize}

\subsection{Loss Function}

In deep learning, optimizing a model's parameters is guided by stochastic gradient descent (SGD) with a loss function. The goal of the SGD is to make the result of the loss function as small as possible.

Generally, the Manhattan distance (L1 norm) and Euclidean distance (L2 norm) are two commonly used loss functions to measure the difference between two series of path points. Given the ground truth positions $(x_t, y_t)$ and the predicted positions $(x_t', y_t')$ where $x$ and $y$ are the horizontal and vertical coordinates and $t = (1, 2, 3, ..., m)$, the L1 and L2 loss function are defined as Equations~\ref{L1} and~\ref{L2},
\begin{equation}
\begin{split}
    l = \frac{1}{2m} \sum_{i=1}^{m} (|{x_t-x_t'}| + |{y_t-y_t'}|),
\end{split}
\label{L1}
\end{equation}

\begin{equation}
\begin{split}
    l = \frac{1}{2m} \sum_{i=1}^{m} [({x_t-x_t'})^2 + ({y_t-y_t'})^2],
\end{split}
\label{L2}
\end{equation}
where $l$ is the loss value. Since vehicles generally drive on the road, the vertical coordinate $z$ is ignored here to simplify the description.

\subsection{Coordinate Transformation}
In practice, the global coordinates $x$ and $y$ have no fixed range. If we directly use the original coordinate values to train a deep neural network, it will be difficult to converge, and the model will lose generalization. Therefore, we need to convert the global coordinates into relative coordinates by Equation~\ref{relative} where $(x_0,y_0)$ is the current position.
\begin{equation}
\begin{split}
    x_t = x_t-x_0, \quad y_t = y_t-y_0.
\end{split}
\label{relative}
\end{equation}

\begin{table*}[ht]
\caption{Coordinate Conversion Comparison}
\renewcommand\arraystretch{1.5}
    \centering
    \begin{tabular}{@{\hspace{0pt}}m{2.5cm}<{\centering}@{\hspace{0pt}}|@{\hspace{0pt}}m{1.6cm}<{\centering}@{\hspace{0pt}}|@{\hspace{0pt}}m{1.6cm}<{\centering}@{\hspace{0pt}}|@{\hspace{0pt}}m{3cm}<{\centering}@{\hspace{0pt}}|@{\hspace{0pt}}m{3.8cm}<{\centering}@{\hspace{0pt}}|@{\hspace{0pt}}m{1.6cm}<{\centering}@{\hspace{0pt}}|@{\hspace{0pt}}m{3.8cm}<{\centering}@{\hspace{0pt}}}
    \hline
    \hline
    &\multicolumn{6}{c}{\textbf{Original Coordinate}} \\
    \hline
    & Current & \multicolumn{5}{c}{Future}  \\
    \cline{2-7}
     & $x_0$ & $x_1$ & $x_2$ & $x_3$ & ...& $x_m$\\
    & $y_0$ & $y_1$ & $y_2$ & $y_3$ & ...& $y_m$\\
    \hline
    \hline
    \textbf{Conversion Method}&& \multicolumn{4}{c}{\textbf{Model’s Target} (Ground Truth)}\\
    \hline
    && $\Delta x_1$ & $\Delta x_2$ & $\Delta x_3$ & ...& $\Delta x_m$\\
    && $\Delta y_1$ & $\Delta y_2$ & $\Delta y_3$ & ...& $\Delta y_m$\\
    \hline
    \multirow{2}{*}{Equation~\ref{relative}} && $x_1-x_0$ & $x_2-x_0$ & $x_3-x_0$ &...& $x_m-x_0$\\
    && $y_1-y_0$ & $y_2-y_0$ & $y_3-y_0$ &...& $y_m-y_0$\\
    \hline
    \multirow{2}{*}{Equation~\ref{delta}} && $x_1-x_0$ & $x_2-x_1$ & $x_3-x_2$ &...& $x_m-x_{m-1}$\\
    && $y_1-y_0$ & $y_2-y_1$ & $y_3-y_2$ &...& $y_m-y_{m-1}$\\
    \hline
    \multirow{2}{*}{Equation~\ref{delta1}} & & $x_1 - x_0$ & $x_2-(x_0+\Delta x_1')$ & $x_3-(x_0+\Delta x_1'+\Delta x_2')$ & ... & $x_m-(x_0+\sum_{i=1}^{m-1} \Delta x_i')$\\
    & & $y_1 - y_0$ & $y_2-(y_0+\Delta y_1')$ & $y_3-(y_0+\Delta y_1'+\Delta y_2')$ & ... & $y_m-(y_0+\sum_{i=1}^{m-1} \Delta y_i'))$\\
    \hline
    \hline
    && \multicolumn{4}{c}{\textbf{Model's Output}} \\
    \hline
    && $\Delta x_1'$ & $\Delta x_2'$ & $\Delta x_3'$ & ... & $\Delta x_m'$\\
    && $\Delta y_1'$ & $\Delta y_2'$ & $\Delta y_3'$ & ... & $\Delta y_m'$\\
    \hline
    \hline
    \end{tabular}
    \label{conversion}
\end{table*}

In this work, we further convert the relative coordinates that can make the performance of the model better. First, we try to convert each coordinate into an increment relative to its previous coordinate, as shown in Equation~\ref{delta}.
\begin{equation}
    \Delta x_t =  x_t - x_{t-1},
    \quad
    \Delta y_t = y_t - y_{t-1}.
    \label{delta}
\end{equation}

In training, the model should output $\{(\Delta x_t', \Delta y_t') | t = (1, 2, 3, ..., m)\}$ and use Equation~\ref{L1} or Equation~\ref{L2}
to compute the loss between it and $\{(\Delta x_t, \Delta y_t) | t = (1, 2, 3, ..., m)\}$. However, in testing, we can not just simply generate the predicted path points $\{(x_t', y_t') | t = (1, 2, 3, ..., m)\}$ by Equation~\ref{dedelta}
\begin{equation}
    x_t' =  x_{t-1} + \Delta x_t',
    \quad
    y_t' = y_{t-1} + \Delta y_t',
    \label{dedelta}
\end{equation}
because $x_{t-1}$ and $y_{t-1}$ are unknown in testing. The correct calculation method is 
\begin{equation}
\begin{split}
    x_t' = x_0 + \sum_{i=1}^{t} \Delta x_i', \quad y_t' = y_0 + \sum_{i=1}^{t} \Delta y_i'.
\end{split}
\label{recovery}
\end{equation}

Now, we can find that, except for the $(\Delta x_1', \Delta y_1')$, in training, the target of $(\Delta x_t', \Delta y_t')$ is $(x_t - x_{t-1}, y_t - y_{t-1})$, but in testing, the target is $[x_t - (x_0 + \sum_{i=1}^{t-1} \Delta x_i'), y_t - (y_0 + \sum_{i=1}^{t-1} \Delta y_i')]$. Therefore, in order to achieve better results in testing, we convert the Equation~\ref{delta} into Equation~\ref{delta1},
\begin{equation}
\begin{split}
    \Delta x_t = 
    \begin{cases} 
        x_1 - x_0 & \text{if } t = 1, \\ 
        x_t - (x_0 + \sum_{i=1}^{t} \Delta x_i') & \text{otherwise}, 
    \end{cases}
    \\ 
    \Delta y_t = 
    \begin{cases} 
        y_1 - y_0 & \text{if } t = 1, \\ 
        y_t - (y_0 + \sum_{i=1}^{t} \Delta y_i') & \text{otherwise},
    \end{cases}
\label{delta1}
\end{split}
\end{equation}
which means in training iterations, the ground truth needs to be dynamically adjusted according to the output of the model.

In order to visually compare different conversion methods, we list the results of each conversion method in Table~\ref{conversion}.

\section{Experiment}

\begin{figure*}[htb]%
    \centering%
    %
    \includegraphics[width=0.49\textwidth]{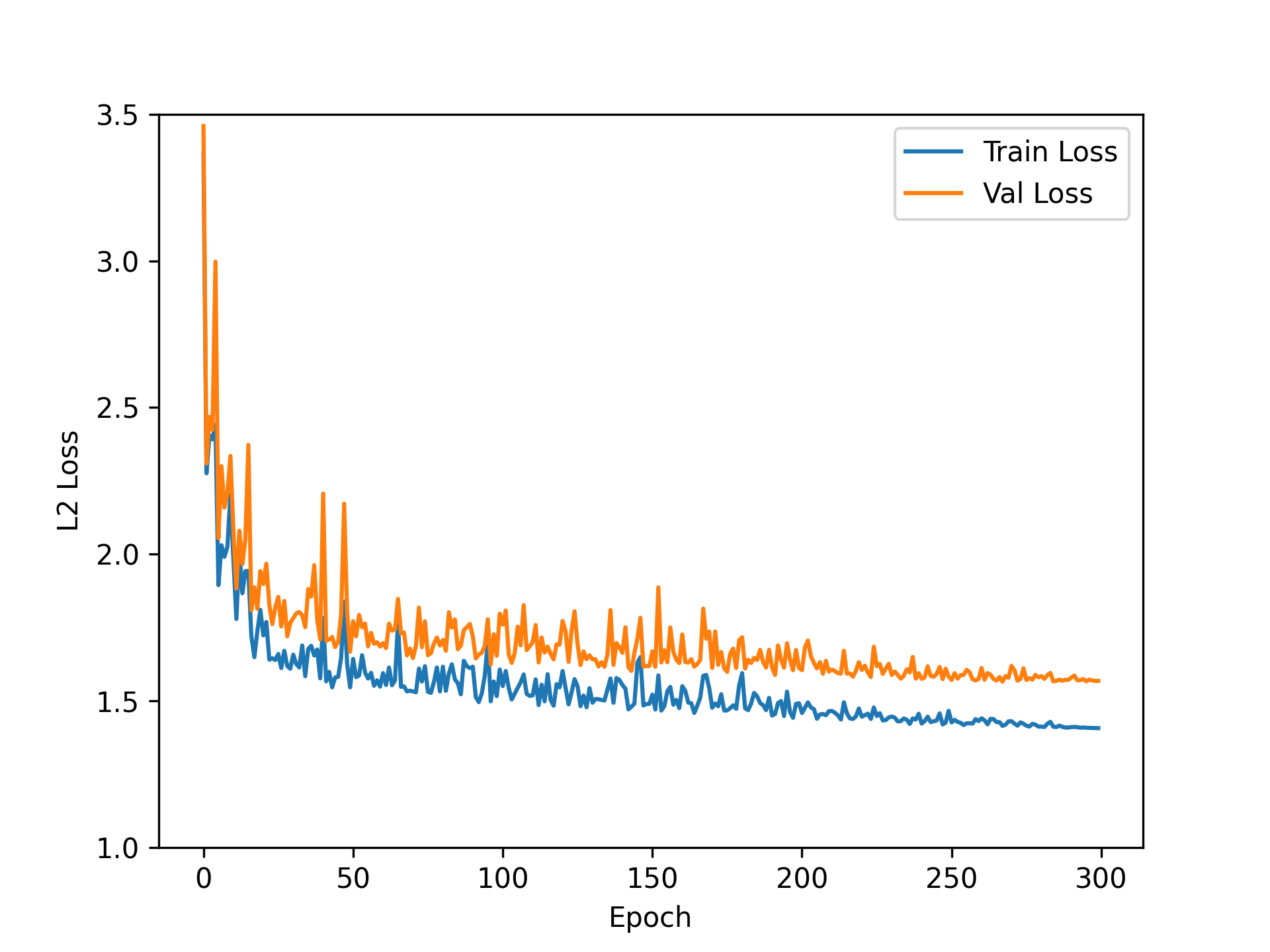}%
    \
    %
    \includegraphics[width=0.49\textwidth]{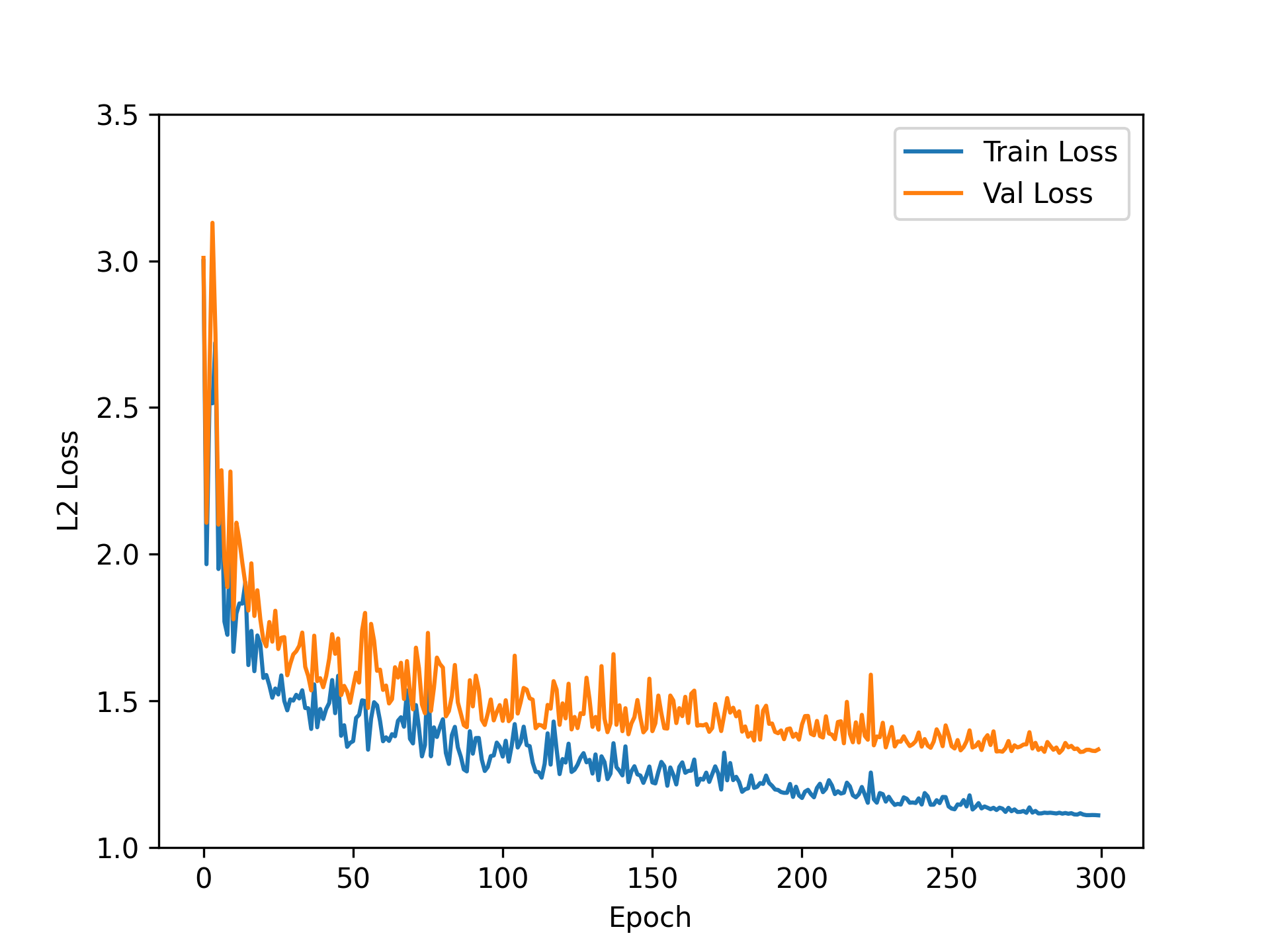}\par%

    \caption{Loss Comparison}%
    \label{loss}%
\end{figure*}

In this section, we train an end-to-end path planning model on nuScenes~\cite{nuscenes2019} dataset. In the training phase, we use $\frac{1}{2m} \sum_{i=1}^{m} [({\Delta x_t-\Delta x_t'})^2 + ({\Delta y_t-\Delta y_t'})^2]$
to guide the network optimization. For comparison, we recover the $(\Delta x_t', \Delta y_t')$ to $(x_t, y_t)$ to calculate both training loss and validation loss. Figure~\ref{loss} shows the change of loss for two coordinate transform methods. The left one uses the method of Equation~\ref{relative}, and the right one uses our method of Equation~\ref{delta1}. Compared with the other method, our method reduces the loss by 15\%.

\section{Conclusion}
This paper presented the Residual Chain Loss method, a novel approach designed to address the challenges of path planning and behavior cloning for autonomous driving system. Our model outputs are formulated as relative coordinates between adjacent path points in the sequence. This adjustment is accomplished by modifying the ground truth to represent the Manhattan distance between the expert coordinates and the sum of the initial point coordinates and the cumulative pre-predicted outputs, in order to achieve dynamic loss adjustment. Thus our method effectively mitigates the issues of covariate shift and enhances model performance without increasing computational cost. Our experiments on the nuScene dataset validate the approach's compatibility with end-to-end path planning models, indicating substantial improvements over traditional relative coordinates methods. The results affirm the potential of Residual Chain Loss to advance the development of more efficient and reliable autonomous driving systems. Future work will explore its 
\section{Acknowledgement}

\bibliographystyle{ieeetr}
\bibliography{references}

\vfill

\end{document}